%% file: acl2023.tex
\pdfoutput=1

\documentclass[11pt]{article}

\usepackage{ACL2023}

\usepackage{times}
\usepackage{latexsym}

\usepackage[T1]{fontenc}

\usepackage[utf8]{inputenc}

\usepackage{microtype}

\usepackage{inconsolata}

\usepackage{graphicx}
\usepackage{hyperref}

\usepackage[boxruled,linesnumbered,vlined,inoutnumbered]{algorithm2e}
\SetKwInOut{Parameter}{Parameters}
\usepackage{epsfig}
\usepackage{amsmath}
\usepackage{amssymb}
\usepackage{float}
\usepackage{multirow}
\usepackage{hhline}
\usepackage{makecell}
\usepackage{tabularx}
\usepackage{bbm}

\usepackage{caption}
\usepackage{booktabs}
\usepackage{bbold}
\usepackage{multirow}
\usepackage{array, makecell} %
\usepackage{dsfont}
\usepackage{listings}
\usepackage{lipsum}
\usepackage{graphicx}
\usepackage{soul}

\input{init}
\input{dfn}
\input{math_commands}

\definecolor{coolgrey}{rgb}{0.55, 0.57, 0.67}

\title{Learning from a Friend: Improving Event Extraction via Self-Training with Feedback from Abstract Meaning Representation}

\author{Zhiyang Xu\textsuperscript{${\spadesuit}$} \quad Jay-Yoon Lee\textsuperscript{$\clubsuit$}\footnotemark[1] \quad Lifu Huang$^{\spadesuit}$\thanks{\hspace{2mm}corresponding authors} \\
  $^{\spadesuit}$Department of Computer Science, Virginia Tech \\
  $^\clubsuit$Graduate School of Data Science,  Seoul National University\\
  \texttt{\{zhiyangx,lifuh\}@vt.edu} \\ 
  {lee.jayyoon@snu.ac.kr}\\}

\begin{document}
\maketitle
\input{0abstract}
\input{1introduction}
\input{3approach}

\input{4experiment}

\input{5discussion}

\input{6relatedwork}

\input{7conclusion}
\input{9limitation}
\input{Acknowledgments}
\bibliography{anthology,custom}
\bibliographystyle{acl_natbib}

\appendix

\input{8appendix}

\label{sec:appendix}

\end{document}

%% file: init.tex
\usepackage{graphicx}               
\usepackage{tabularx}               
\usepackage{booktabs}
\usepackage{amsmath}
\usepackage{stmaryrd}
\usepackage{xcolor}
\usepackage{soul}

\newcolumntype{C}{>{\centering\arraybackslash}X}
\usepackage{multirow}               
\usepackage{diagbox}                
\usepackage{hhline}                 
\usepackage{color}                  
\usepackage{amsmath}                
\usepackage{amssymb}                
\usepackage{mathtools}              
\usepackage{enumitem}               

\usepackage{subfigure}
\usepackage{booktabs}
\usepackage{extarrows}
\usepackage{makecell}

\usepackage{soul}

\definecolor{RoseQuartzBg}{HTML}{F7CAC9}
\definecolor{RoseQuartz}{HTML}{F5A798}
\definecolor{Serenity}{HTML}{92A8D1}
\definecolor{OrangeRed}{rgb}{1.0, 0.27, 0.0}
\definecolor{Red}{rgb}{1.0, 0.0, 0.0}
\definecolor{Turquoise}{HTML}{0F4C81}
\usepackage{xparse}
\NewDocumentCommand{\lifu}{ mO{} }{\textcolor{OrangeRed}{\textsuperscript{\textit{Lifu}}\textsf{\textbf{\small[#1]}}}}
\NewDocumentCommand{\minqian}{ mO{} }{\textcolor{blue}{\textsuperscript{\textit{Minqian}}\textsf{\textbf{\small[#1]}}}}
\NewDocumentCommand{\zhiyang}{ mO{} }{\textcolor{Serenity}{\textsuperscript{\textit{Zhiyang}}\textsf{\textbf{\small[#1]}}}}

\NewDocumentCommand{\jy}{ mO{} }{\textcolor{blue}{\textsuperscript{\textit{jy}}\textsf{\textbf{\small[#1]}}}}

%% file: dfn.tex
\newcommand{\stgg}{\textsc{Stf}}
\newcommand{\stggamr}{${\textsc{Stf}_\textsc{AMR}}$}
\newcommand{\stggtext}{${\textsc{Stf}_\textsc{w/o\_AMR}}$}
\newcommand{\selftraining}{\textit{self-training}}

\definecolor{applegreen}{rgb}{0.55, 0.71, 0.0}

\newcommand{\ace}{ACE05-E}
\newcommand{\aceplus}{$\text{ACE05-E}^+$}
\newcommand{\ere}{ERE-EN}

\newcommand{\vvv}[1]{\ensuremath{\mathbf{\lowercase{#1}}}} 
\newcommand{\mat}[1]{\ensuremath{\mathbf{\uppercase{#1}}}} 

%% file: math_commands.tex
\usepackage{amsmath,amsfonts,bm}

\def\eqref#1{equation~\ref{#1}}

\def\1{\bm{1}}

\DeclareMathAlphabet{\mathsfit}{\encodingdefault}{\sfdefault}{m}{sl}
\SetMathAlphabet{\mathsfit}{bold}{\encodingdefault}{\sfdefault}{bx}{n}



%% file: 0abstract.tex
\begin{abstract}
Data scarcity has been the main factor that hinders the progress of event extraction. To overcome this issue, we propose a \textbf{S}elf-\textbf{T}raining with \textbf{F}eedback (\stgg{}) framework that leverages the large-scale unlabeled data and acquires feedback for each new event prediction from the unlabeled data by comparing it to the Abstract Meaning Representation (AMR) graph of the same sentence. Specifically, \stgg{} consists of (1) a base event extraction model trained on existing event annotations and then applied to large-scale unlabeled corpora to predict new event mentions as pseudo training samples, and (2) a novel scoring model that takes in each new predicted event trigger, an argument, its argument role, as well as their paths in the AMR graph to estimate a compatibility score indicating the correctness of the pseudo label. The compatibility scores further act as feedback to encourage or discourage the model learning on the pseudo labels during self-training. Experimental results on three benchmark datasets, including \ace{}, \aceplus{}, and ERE, demonstrate the effectiveness of the \stgg{} framework on event extraction, especially event argument extraction, with significant performance gain over the base event extraction models and strong baselines. Our experimental analysis further shows that \stgg{} is a generic framework as it can be applied to improve most, if not all, event extraction models by leveraging large-scale unlabeled data, even when high-quality AMR graph annotations are not available.\footnote{The source code and model checkpoints are publicly available at \url{https://github.com/VT-NLP/Event_Extraction_with_Self_Training}.}

\end{abstract}

%% file: 1introduction.tex
\section{Introduction}

Event extraction (EE), which aims to identify and classify event triggers and arguments, has been a long-stand challenging problem in natural language processing. Despite the large performance leap brought by advances in deep learning, recent studies \cite{deng2021ontoed, CLEVE} have shown that the data scarcity of existing event annotations has been the major issue that hinders the progress of EE. For example, in ACE-05\footnote{\url{https://www.ldc.upenn.edu/collaborations/past-projects/ace}}, one of the most popular event extraction benchmark datasets, 10 of the 33 event types have less than 80 annotations. 
However, creating event annotations is extremely expensive and time-consuming, e.g., it takes several linguists over one year to annotate 500 documents with about 5000 event mentions for ACE-05. 

To overcome the data scarcity issue of EE, previous studies~\cite{chen2009can,liao2011can,ferguson2018semi} develop self-training methods that allow the trained EE model to learn further by regarding its own predictions on large-scale unlabeled corpora as pseudo labels.  
However, simply adding the high-confidence event predictions to the training set inevitably introduces noises~\cite{gradual_drift_ner, confirmation_bias, mentor_net}, especially given that the current state-of-the-art performance of event argument extraction is still less than 60\% F-score. To tackle this challenge, we introduce a \textbf{S}elf-\textbf{T}raining with \textbf{F}eedback framework, named \stgg{}, which consists of an \textit{event extraction model} that is firstly trained on the existing event annotations and then continually updated on the unlabeled corpus with self-training, and a \textit{scoring model} that is to evaluate the correctness of the new event predictions (pseudo labels) from the unlabeled corpus, and the scores further act as feedback to encourage or discourage the learning of the event extraction model on the pseudo labels during self-training, inspired by the REINFORCE algorithms~\cite{reinforce}.


Specifically, the event extraction model of our \stgg{} framework can be based on any state-of-the-art architecture. In this paper, we choose OneIE~\cite{OneIE} and AMR-IE~\cite{AMR_IE}, due to their superior performance and publicly available source code. The scoring model leverages the Abstract Meaning Representation (AMR)~\cite{amr} which has been proven to be able to provide rich semantic and structural signals to map AMR structures to event predictions~\cite{huang2016liberal,huang2018zero,CLEVE} and thus their compatibility can indicate the correctness of each event prediction. 
The scoring model is a self-attention network that takes in a predicted event trigger, a candidate argument and its argument role, as well as their path in the AMR graph of the whole sentence, and computes a score ranging in [-1, 1] based on the compatibility between the AMR and the predicted event structure: -1 means \textit{incompatible}, 1 means \textit{compatible}, and 0 means \textit{uncertain}. Inspired by the REINFORCE algorithm~\cite{reinforce}, we multiply the compatibility scores and the gradient of the EE model computed on the pseudo event labels during self-training, so as to (1) encourage the event extraction model to follow the gradient and hence maximize the likelihood of the pseudo label when it is compatible with the AMR structure; (2) negate the gradient and minimize the likelihood of the pseudo label when it is incompatible with the AMR structure; and (3) reduce the magnitude of the gradient when the scoring model is uncertain about the correctness of the pseudo label.

We take AMR 3.0\footnote{https://catalog.ldc.upenn.edu/LDC2020T02.} and part of the New York Times (NYT) 2004 corpus\footnote{https://catalog.ldc.upenn.edu/LDC2008T19} as additional unlabeled corpora to enhance the event extraction model with \stgg{}, and evaluate the event extraction performance on three public benchmark datasets: \ace{}\footnote{https://catalog.ldc.upenn.edu/LDC2006T06}, \aceplus{}\footnote{https://catalog.ldc.upenn.edu/LDC2006T06}, and \ere{}\footnote{Deep Exploration and Filtering of
Test (DEFT) program.}. The experimental results demonstrate that: (1) the vanilla \selftraining{} barely improves event extraction due to the noise introduced by the pseudo examples, while the proposed \stgg{} framework leverages the compatibility scores from the scoring model as feedback and thus makes more robust and efficient use of the pseudo labels; (2) \stgg{} is a generic framework and can be applied to improve most, if not all, of the event extraction models optimized by gradient descent algorithm and achieves significant improvement over the base event extraction models and strong baselines on event argument extraction on the three public benchmark datasets; (3) By exploiting different unlabeled corpora with gold or system-based AMR parsing, \stgg{} always improves the base event extraction models, demonstrating that it can work with various qualities of AMR parsing. 
Notably, different from previous studies~\cite{huang2018zero,AMR_IE,CLEVE} that require high-quality AMR graphs as input to the model during both training and inference, \stgg{} does not require any AMR graphs during inference, making it more computationally efficient and free from the potential errors propagated from AMR parsing.

%% file: 3approach.tex
\section{\stgg{} for Event Extraction}
The event extraction task consists of three subtasks: event detection, argument identification and argument role classification. Given an input sentence $W=[w_1, w_2, ... ,w_N]$, \textit{event detection} aims to identify the span of an event trigger $\tau_i$ in $W$ and assign a label $l_{\tau_i}\in\mathcal{T}$ where $\mathcal{T}$ denotes the set of target event types.
\textit{Argument identification} aims to find the span of an argument $\varepsilon_j$ in $W$, and \textit{argument role classification} further predicts a role $\alpha_{ij}\in\mathcal{A}$ that the argument $\varepsilon_j$ plays in an event $\tau_i$ given the set of target argument roles $\mathcal{A}$. 

\begin{figure*}[ht!]
    \centering
    \includegraphics[width=\textwidth]{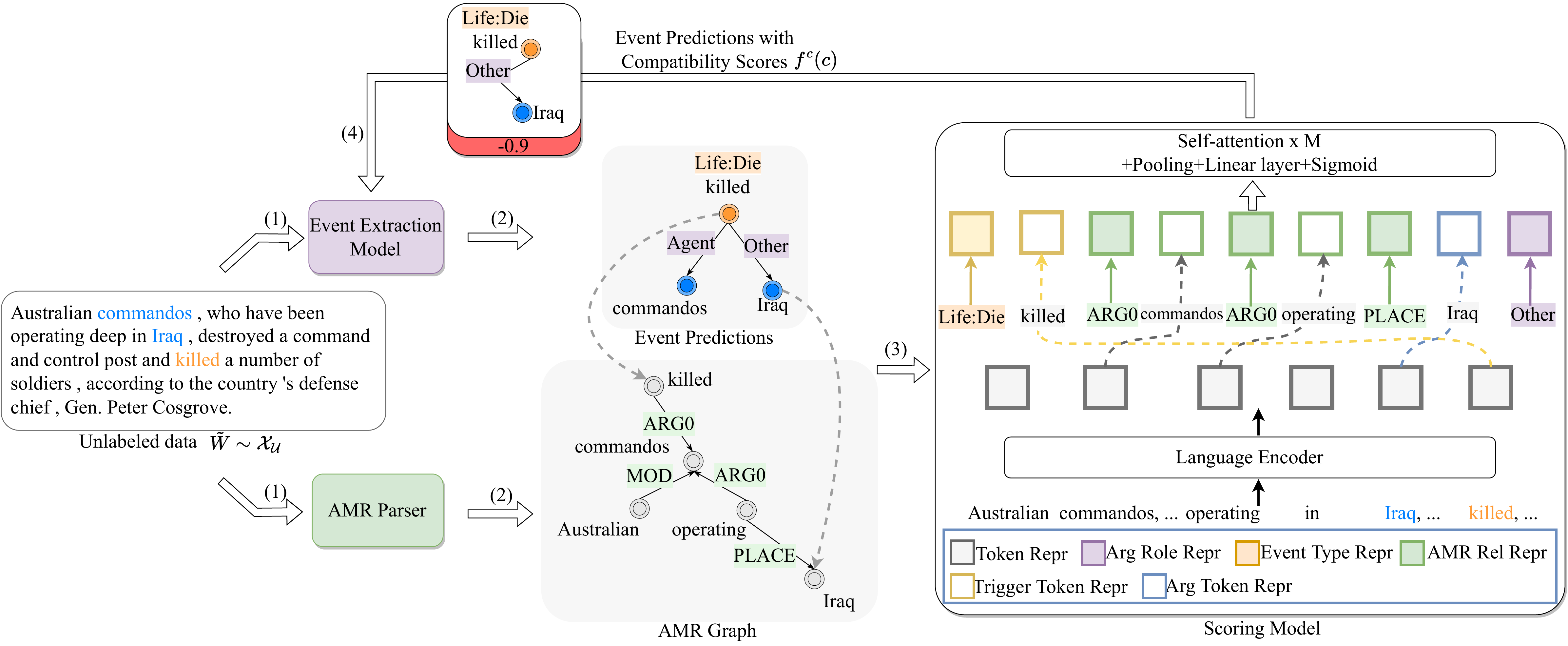}
    \vspace{-5mm}
    \caption{ The overall framework of \stgg{} (We omit the first stage of \stgg{}.). Given an unlabeled sentence $\tilde{W}$: (1) run an event extraction model to compute event predictions and an AMR parser to parse it into an AMR graph; (2) map the predicted trigger and argument to corresponding nodes in the AMR graph, find their AMR path and combine it with the predicted event type and argument role into a new sequence (Life:Die, killed, ARG0, commandos, ARG0, PLACE, Iraq, Other); 
    (3) feed the sequence into the scoring model to compute a compatibility score; (4) leverage the pseudo label and compatibility score to further update the event extraction model. 
    }
    \vspace{-4mm}
    \label{fig:model_arch}
\end{figure*}

Figure~\ref{fig:model_arch} shows the overview of our \stgg{} framework which consists of two training stages. 
At the first stage, a \textbf{base event extraction model} (Section~\ref{subsec:base-event}) is trained on a labeled dataset.
At the second stage, we apply the trained event extraction model to an unlabeled corpus to predict new event mentions. Instead of directly taking the new event predictions as pseudo training examples like the vanilla \selftraining{}, we propose a novel \textbf{scoring model} (Section~\ref{sec:score_model_method}) to estimate the correctness of each event prediction by measuring its compatibility to the corresponding AMR graph, and then take both event predictions and their compatibility scores to continue to train the base event extraction model while the scores update the gradient computed on pseudo labels (Section~\ref{subsec:stf}). After the training of the second stage, we get a new event extraction model and evaluate it on the test set. 

\subsection{Base Event Extraction Model}
\label{subsec:base-event}
Our proposed framework can be applied to most, if not all, event extraction models. We select OneIE~\cite{OneIE} and AMR-IE~\cite{AMR_IE} as base models given their state-of-the-art performance on the event extraction task and publicly available source code. Next, we briefly describe the common architectures in the two models and refer readers to the original papers for more details.
OneIE and AMR-IE perform event extraction in four\footnote{We only focus on event extraction task and thus omit the description of relation extraction.} steps. 
\textbf{First}, a language model encoder~\cite{bert,roberta} computes the contextual representations $\mat{W}$ for an input sentence $W$. \textbf{Second}, two identification layers take in the contextual representations $\mat{W}$. One identifies the spans of event triggers and the other identifies the spans of arguments (i.e., entities). Both of them are based on a linear classification layer followed by a CRF layer~\cite{crf} to capture the dependencies between predicted tags. 
They are optimized by minimizing the negative log-likelihood of the gold-standard tag path, which is denoted as 
$\mathbf{L}^\text{Tri\_I}$ 
and $\mathbf{L}^\text{Arg\_I}$ for trigger and argument identification, respectively.
\textbf{Third}, for each trigger or argument candidate, we compute its representation by averaging the token representations within the whole identified span. Each trigger representation is fed into a classification layer to predict its type by minimizing the cross-entropy classification loss $\mathbf{L}^\text{Tri\_C}$. Each pair of trigger and argument representations are concatenated and fed into another classification layer to predict the argument role, which is also optimized by the cross-entropy loss $\mathbf{L}^\text{Arg\_C}$. \textbf{Finally}, both OneIE and AMR-IE learn an additional global feature vector to capture the interactions across sub-tasks (e.g., a \textit{LOC} entity is impossible to be the \textit{Attacker} of an \textit{Attack} event) and instances (e.g., the \textit{Defendant} of a \textit{Sentence} event can also be an \textit{Agent} of a \textit{Die} event). During training, a global feature score is computed for the predicted information graph and the gold annotation, respectively, from their global feature vectors. The training objective is to minimize the gap between these two global feature scores, denoted as $\mathbf{L}^{G}$. Thus, the overall loss for the base event extraction model is:
\begin{align*}
\mathbf{L}^\text{E} = \mathbf{L}^\text{Tri\_I} + \mathbf{L}^\text{Arg\_I} + \mathbf{L}^\text{Tri\_C} + \mathbf{L}^\text{Arg\_C} + \mathbf{L}^{G},
\label{eq:base_obj}
\end{align*}

As the first stage of our \stgg{} framework, we optimize the base event extraction model on labeled event mentions $\mathcal{X}_L$ based on $\mathbf{L}^\text{E}$ and the trained model will later be used to predict new event mentions for self-training.

\subsection{Scoring Model} \label{sec:score_model_method}


At the second stage of \stgg{}, we aim to further improve the event extraction model by taking the event mentions predicted from an external unlabeled corpus $\mathcal{X}_u$ as pseudo samples for self-training. 
To avoid the noise contained in the pseudo samples, we propose a scoring model that can evaluate the correctness of each event prediction.
Our scoring model takes AMR graph as a reference motivated by the observation that \textit{an event structure usually shares similar semantics and network topology as the AMR graph of the same sentence, thus their compatibility can be used to measure the correctness of each event structure}. 
This observation has also been discussed and shown effective in previous studies~\cite{amr_ie_sudha2017,huang2018zero,AMR_IE}. 
However, previous studies directly take AMR graphs as input to the extraction model and thus require AMR graphs during both training and inference, making their performance highly dependent on the quality of AMR parsing. Different from them, our proposed \stgg{} only takes AMR graphs during reference to measure the correctness of event predictions during self-training, making it free from the potential errors propagation from AMR parsing during inference.




Given a sentence $\tilde{W}\in\mathcal{X}_u$ from the unlabeled corpus and a predicted trigger $\tilde{\tau}_i$ and its argument $\tilde{\varepsilon}_j$ from $\tilde{W}$, we aim to estimate a correctness score for each pair of the trigger and argument prediction 
based on its compatibility with their path in the AMR graph\footnote{Comparing with the whole AMR graph, the path of the trigger and argument in the AMR graph shows more improvement for the scoring model.}. Thus, we first apply the state-of-the-art AMR parsing tool~\cite{transformer_amr_parser} to generate an AMR graph for $\tilde{W}$: $G=(V,E)$, $E= \{(v_i, e_{ij}, v_j)|e_{ij}\in\mathcal{R})\}$. 
We follow ~\cite{huang2016liberal,AMR_IE} and group the original set of AMR relations into $19$ categories\footnote{The details of AMR relation categories are shown in Appendix~\ref{sec:AMR_rel_groups}.}, thus $e_{ij}$ denotes a particular relation category and $\mathcal{R}$ denotes the set of AMR relation categories. 
Then, we identify the $v_i$, $v_j$ from AMR graph $G$
as the corresponding node of  $\tilde{\tau}_i$, $\tilde{\varepsilon}_j$,  
by node alignment following~\citet{AMR_IE}. Then, we utilize the Breadth First Search to find the shortest path $p_{i,j}$ that connects and includes, $v_i$ and $v_j$ in $G$. If there is no path between $v_i$ and $v_j$, we add a new edge to connect them and assign \textit{other} as the relation. 




Given a predicted trigger $\tilde{\tau}_i$ and its type $\tilde{l}_{\tau_i}$, a predicted argument $\tilde{\varepsilon}_j$ and its argument role $\tilde{\alpha}_{ij}$, the scoring model estimates their correctness by taking $[\tilde{l}_{\tau_i}$, $p_{ij}$, $\tilde{\alpha}_{ij}]$ as input and outputs a compatibility score. As Figure~\ref{fig:model_arch} shows, it consists of a language model encoder \cite{bert,roberta} to encode the sentence $\tilde{W}$ and obtain the contextual representations for the tokens\footnote{If a token is split into multiple subtokens, we average the representations of all subtokens to obtain an overall token representation.}, which are then used to initialize the representation of each node in $p_{ij}$ based on the alignment between the input tokens and the nodes in AMR graph following~\citet{AMR_IE}. 
We draw edge representations from the AMR relation embedding matrix $\mat{E}^{rel}$ and combine them with node representations to form $\mat{h}_{p_{ij}}$, a representation for path $p_{ij}$. We also get an event type representation $\vvv{h}_{\tau_i}$ for $\tilde{l}_{\tau_i}$ from the event-type embedding matrix $\mat{E}^{tri}$ and an argument role representation $\vvv{h}_{\alpha_{ij}}$ for $\tilde{\alpha}_{ij}$
from the argument role embedding matrix $\mat{E}^{arg}$. Here, $\mat{E}^{rel}$, $\mat{E}^{tri}$, and $\mat{E}^{arg}$ are all randomly initialized and will be optimized during training. Finally, we obtain the initial representations $\mat{H}^{init}_{ij}=[\vvv{h}_{\tau_i},\mat{h}_{p_{ij}},\vvv{h}_{\alpha_{ij}}]$ for the sequence $[\tilde{l}_{\tau_i}, p_{ij}, \tilde{\alpha}_{ij}]$.

To estimate the compatibility between the event trigger and argument prediction and their path in the AMR graph, we  apply multi-layer \textit{self-attention}~\cite{transformer} over the joint representation of the AMR path and the event prediction $\mat{H}^{init}_{ij}$ to learn better contextual representations for the sequence $[\tilde{l}_{\tau_i}, p_{ij}, \tilde{\alpha}_{ij}]$ and we add the position embedding $\mat{E}^{pos}$ to $\mat{H}^{init}_{ij}$ before feed it into the self-attention layers:
\begin{align*}
 \mat{H}^{final}_{ij}= \textit{self-attention}(\mat{H}^{init}_{ij})\times M,
\end{align*}
where $M$ denotes the number of attention layers.

Finally, we compute an overall vector representation $\hat{\mat{H}}^{final}_{ij}$ from $\mat{H}^{final}_{ij}$ via average-pooling and feed it into a \textit{linear-layer} and a \textit{Sigmoid} function to compute a probability $c_{ij}$, indicating the correctness of the predicted event trigger and argument. We optimize the scoring model based on the binary cross-entropy objective:
\begin{align*}
 \mathbf{L}^\text{Score}= \text{BCE}\left(y_{ij}, c_{ij}; \psi\right),
\end{align*}
where $y_{ij}\in(0,1)$ is a binary label that indicates the argument role is correct ($y_{ij}=1$) or not ($y_{ij}=0$)\footnote{We don't not consider the cases where the trigger labels are incorrect, since by observation the semantics and structure of AMR graphs are more related to the argument role types between event triggers and their arguments.}, and $\psi$ is the parameters of the scoring model. During training, we have gold triggers and arguments as positive training instances and we swap the argument roles in positive training instances with randomly sampled incorrect labels to create negative training instances. After training the scoring model, we will fix its parameters and apply it to self-training. 

\subsection{Self-Training with Feedback}
\label{subsec:stf}
To improve the base event extraction model with \selftraining{}, we take the new event predictions ($\tilde{\tau}_i$, $\tilde{l}_{\tau_i}$, $\tilde{\varepsilon}_j$, $\tilde{\alpha}_{ij}$) from the unlabeled corpus $\mathcal{X}_u$ as pseudo samples to further train the event extraction model. 
The gradients of the event extraction model on each pseudo sample is computed as:
\begin{align*}
g^{st}_{ij} &= \nabla_\theta \mathbf{L}^\text{E}\left(\tilde{W}, (\tilde{\tau}_i, \tilde{l}_{\tau_i}, \tilde{a}_{ij}, \tilde{\varepsilon}_j);\theta\right)
\end{align*}
where $\theta$ denotes the parameters of the event extraction model. 
Note that there can be multiple event predictions in one sentence.


Due to the prediction errors of the pseudo labels, simply following the gradients $g^{st}_{ij}$ computed on the pseudo labels can hurt model's performance. Thus, we utilize the correctness score $c_{ij}$ predicted by the scoring model 
to update the gradients, based on the motivation that: (1) if an event prediction is compatible with the AMR structure, it's likely to be correct and we should encourage the model learning on the pseudo label; (2) on the other side, if an event prediction is incompatible with its AMR structure, it's likely incorrect and we should discourage the model learning on the pseudo label; (3) if the scoring model is uncertain about the correctness of the event prediction, we should reduce the magnitude of the gradients learned from the pseudo label. Motivated by this, we first design a transformation function $f^c$ to project the correctness score $c_{ij}\in[0, 1]$ into a range $[-1,1]$ where -1 (or $c_{ij}=0$) indicates incompatible, 1 (or $c_{ij}=1$) means compatible, and 0 (or $c_{ij}=0.5$) means uncertain.
 Here, $f^c$ is based on a linear mapping:
\begin{equation*}
 f^c(c_{ij})=2\times c_{ij} -1   
\end{equation*}


We then apply the compatibility scores as feedback to update the gradients of the event extraction model on each pseudo sample during self-training:
\begin{align*}
\mathbf{L}^\text{\stgg{}} = \sum_{i,j}f^c(c_{ij}) \cdot \mathbf{L}^\text{E}\left(\tilde{W}, (\tilde{\tau}_i, \tilde{l}_{\tau_i}, \tilde{a}_{ij}, \tilde{\varepsilon}_j);\theta\right)
\end{align*}

To improve the efficiency of \selftraining{}, we update the event extraction model on every minibatch, and to avoid the model diverging, we combine the supervised training and self-training, so the overall loss for \stgg{} is:
\begin{align*}
\mathbf{L} = \mathbf{L}^\text{E} + \beta\mathbf{L}^\text{\stgg{}}
\end{align*}
where $\beta$ is the combining ratio, $\mathbf{L}^\text{E}$ is computed on the labeled dataset $\mathcal{X}_L$ and $\mathbf{L}^\text{\stgg{}}$ is computed on the pseudo-labeled instances from $\mathcal{X}_u$.

%% file: 4experiment.tex
\section{Experimental Setups}



For evaluation, we consider two base event extraction models: OneIE~\cite{OneIE} and AMR-IE~\cite{AMR_IE} due to their superior performance on event extraction and publicly available source code, and demonstrate the effectiveness of \stgg{} on three benchmark datasets: \ace{}, \aceplus{} and \ere{}, with the same evaluation metrics following previous studies~\cite{DyGIE++,OneIE,AMR_IE,Sijia2022_Query_Extract}\footnote{The detailed statistics of \ace{}, \aceplus{}, and \ere{} are shown in Appendix~\ref{sec:dataset_stat}.}. To show the generalizability of \stgg{}, we explore two unlabeled corpora for self-training: (1) \textbf{AMR 3.0}~\cite{knight2021abstract} which originally contains 55,635 sentences in the training set while each sentence is associated with a manually annotated AMR graph. 
(2) \textbf{New York Times Annotated Corpus} (NYT) contains over 1.8 million articles that were published between 1987 to 2007. We randomly sample 55,635 sentences\footnote{To show the effect of unlabeled dataset vs labeled dataset, we sample the same number of the unlabeled sentences as AMR 3.0} from articles published in 2004. Because NYT dataset does not have AMR annotations, we run a pre-trained AMR parser~\cite{transformer_amr_parser} to generate system AMR parsing.

Besides taking the recent state-of-the-art event extraction studies~\cite{DyGIE++,Du2020_bert_qa_arg,OneIE,FourIE,AMR_IE,lu2021_text2event,Sijia2022_Query_Extract,Hsu2021DEGREE,UIElu2022}\footnote{The scores reported in \cite{nguyen2022graphIE} are not comparable in the table~\ref{tab:main_results}, as their results are not averaged across random seeds. We tried to report their averaged performance by running their model ourselves by contacting the authors, however, their code is publicly unavailable.} as baselines, we also compare our proposed \stgg{} with two other training strategies: (1) vanilla \textit{Self-Training}~\cite{self-training2005} which consists of two stages similar as \stgg{} but in the second stage
takes each new event prediction from the unlabeled data with a probability higher than 0.9 based on the base event extraction model as a pseudo label and combines them with the labeled data to re-train the event extraction model; and (2) \textit{Gradient Imitation Reinforcement Learning} (GradLRE)~\cite{rl_rel}. 
GradLRE encourages the gradients computed on the pseudo-labeled data to imitate the gradients computed on the labeled data by using the cosine distance between the two sources of gradients as a reward to perform policy gradient reinforcement learning~\cite{Sutton1999}. GradLRE showed improvements over other self-training methods on low-resource relation extraction which is a similar task to argument role classification. Appendix~\ref{appendix_training} describes the training details for both baselines and our approach.

%% file: 5discussion.tex
\section{Results and Discussion} \label{sec:result}

\subsection{Evaluation of Scoring Model} \label{sec:score_model}
We first evaluate the performance of the scoring model by measuring how well it distinguishes the correct and incorrect argument role predictions from an event extraction model. Specifically, we compute event predictions by running a fully trained event extraction model (i.e., OneIE or AMR-IE) on the validation and test sets of the three benchmark datasets. Based on the gold event annotations, we create a gold binary label (\textit{correct} or \textit{incorrect}) for each argument role prediction to indicate its correctness. For each event prediction, we pass it along with the corresponding AMR graph of the source sentence into the scoring model. If the correctness\footnote{When the correctness score $c^{ij}>0.5$ computed by the scoring model, the predicted label is \textit{correct}, otherwise, \textit{incorrect}.} 
predicted by the scoring model agrees with the gold binary label, we treat it as a true prediction for scoring model, otherwise, a false prediction.

To examine the impact of leveraging AMR in scoring model performance, we develop a baseline scoring model that shares the same structure with our proposed scoring model except that it does not take an AMR graph as an input. Specifically, the baseline scoring model just takes the event mention (triggers, arguments and argument labels) in order to measure the compatibility score.
The baseline scoring model is essentially an ablation of our scoring model where the AMR path is absent. 
As shown in Table~\ref{tab:score_model}, the performance of our scoring model outperforms the baseline scoring model by 1.4-1.7 F-score on the test sets, demonstrating the effectiveness of AMR graph in characterizing the correctness of each event prediction. 

In Table~\ref{tab:qual_results_scoring_model}, we can observe that the semantics and structure of AMR paths can be easily mapped to argument role types. Sometimes, the even triggers are far from their arguments in plain text, but the AMR paths between them is short and informative. Another observation is that the scoring model tends to assign positive scores to argument roles that are more compatible with the AMR paths, although sometimes the scores for the gold argument roles are not the highest.

\begin{table*}[ht!]
\begin{center}
\resizebox{\textwidth}{!}{%
\begin{tabular}{ |l | l| }
\toprule
\makecell{Tell that to the family of \textcolor{blue}{Margaret Hassan}, the school teacher \\ who was brutally tortured and then \textcolor{orange}{slaughtered} by these same \\ guys, they aren't so bad are they Chris Matthews?} & \makecell{AMR Path: [\textcolor{orange}{slaughtered}, ARG1, teacher, MODIFIER, \textcolor{blue}{Margaret Hassan}] \\ Pred Arg: O; Compatibility Score: \textcolor{red}{-0.99} \\ Gold Arg: Victim; Compatibility Score: \textcolor{applegreen}{0.99}}\\
\midrule
\makecell{It is irritating enough to get \textcolor{orange}{sued} by \textcolor{blue}{Sam Sloan}; imagine how \\ irritating it would be to get BEATEN by him because you have \\done something so egregious that a court is forced to agree with him.} & \makecell{AMR Path: [\textcolor{orange}{sued}, ARG0, \textcolor{blue}{Sam Sloan}] \\ Pred Arg: Adjudicator; Compatibility Score: \textcolor{red}{-0.99} \\ Gold Arg: Plaintiff; Compatibility Score: \textcolor{applegreen}{0.67}} \\
\midrule
  \makecell{\textcolor{orange}{Protests} against the action aimed at toppling Iraqi President Saddam \\ Hussein were held in cities across Libya, Egypt and Lebanon,\\ as well as in \textcolor{blue}{Amman}, Damascus and the Gaza Strip.} & \makecell{AMR Path: [\textcolor{orange}{Protests}, ARG0, held, ARG1, were, PLACE, \textcolor{blue}{Amman}] \\ Pred Arg: O; Compatibility Score: \textcolor{red}{-0.52} \\ Gold Arg: Place; Compatibility Score: \textcolor{applegreen}{0.74}} \\
\midrule
  \makecell{Meanwhile Blair \textcolor{orange}{arrived} in Washington late Wednesday for two \\days of talks with Bush at the Camp David presidential \textcolor{blue}{retreat}.} & \makecell{AMR Path: [\textcolor{orange}{arrived}, OTHER, talks, PLACE, \textcolor{blue}{retreat}] \\ Pred Arg: Destination; Compatibility Score: \textcolor{red}{-0.94} \\ Gold Arg: O; Compatibility Score: \textcolor{applegreen}{0.31}} \\
\bottomrule
\end{tabular}
}
\end{center}
\vspace{-0.4cm}
\caption{Qualitative Results of the compatibility scores.
}
\label{tab:qual_results_scoring_model}
\end{table*}

\begin{table}[ht!]
\begin{center}
\resizebox{0.96\columnwidth}{!}{%
\begin{tabular}{ l | r | r | r | r | r | r }
\toprule
& \multicolumn{2}{c|}{ACE05-E} &
\multicolumn{2}{c|}{ACE05-E+} &
\multicolumn{2}{c}{ERE-EN} \\
\midrule
& Dev & Test & Dev & Test & Dev & Test  \\
\midrule
Scoring w/o AMR & 87.4  & 85.9 &87.9 & 86.9  & 82.8 & 83.1 \\
Scoring w/ AMR & \textbf{88.2} & \textbf{87.4} & \textbf{88.8} & \textbf{88.6} & \textbf{84.4} & \textbf{84.5} \\
\bottomrule
\end{tabular}
}
\end{center}
\vspace{-0.4cm}
\caption{The F-score (\%) of the scoring models on various datasets. Scoring w/o AMR is the baseline scoring model without using AMR path. Scoring w/ AMR is the scoring model we proposed.}
\label{tab:score_model}
\end{table}

\begin{table*}[ht!]
\scriptsize
\begin{center}
\resizebox{0.8\textwidth}{!}{%
\begin{tabular}{l | c  c | c  c | c  c }
\toprule
& \multicolumn{2}{c}{ACE05-E} &
\multicolumn{2}{c}{ACE05-E+} &
\multicolumn{2}{c}{ERE-EN} \\

\midrule
 &  Tri-C  & Arg-C & Tri-C & Arg-C & Tri-C & Arg-C \\
\midrule
DyGIE~\cite{DyGIE++}  & 69.7 & 48.8 & 67.3 & 42.7 & - & -\\
BERT\_QA\_Arg~\cite{Du2020_bert_qa_arg} & 72.4  & 53.3  & 70.6 & 48.3 & 57.0 & 39.2\\
$\text{FourIE}$~\cite{FourIE}  & \underline{\textbf{75.4}} & \textbf{58.0} & 73.3 & \textbf{57.5} & 57.9 & 48.6\\ 
Text2Event~\cite{lu2021_text2event} & 71.9 & 53.8  & 71.8 & 54.4 & 59.4 & 48.3 \\
DEGREE~\cite{Hsu2021DEGREE} & 73.3 & 55.8 & 70.9 & 56.3 & 57.1 & 49.6\\
Query\_Extract~\cite{Sijia2022_Query_Extract} & - & - & \textbf{73.6}  & 55.1 & \textbf{60.4} & \textbf{50.4} \\
UIE~\cite{UIElu2022} & 73.4 & 54.8 & - & - & - & - \\

\midrule
Base OneIE~\cite{OneIE}	& 74.0 & 57.4  & 73.4 & 57.2 & 60.2	& 49.8     \\
+$\text{\selftraining{}}^*$~\cite{self-training2005} & 74.0 &	57.2  	&\underline{\textbf{73.8}}  &	57.3  & 60.1 & 49.4\\
+$\text{GradLRE}^*$~\cite{rl_rel} 	& 74.6	 & 57.4   & 73.5	 & 57.4    & 60.5 & 50.3\\
+${\textsc{\stgg{}}_\textsc{w/o\_AMR}}$ & 74.4 & 57.9 &	\underline{\textbf{73.8}} & 57.6 & 60.4 & 51.0 \\ 
 +${\textsc{\stgg{}}_{\textsc{AMR}}}$ (ours)  &\textbf{75.0} & \underline{\textbf{58.9}} &73.4 &\underline{\textbf{59.0}} &  \underline{\textbf{60.6}} & \underline{\textbf{52.0}} \\
\midrule
Base AMR-IE~\cite{AMR_IE} & 74.4	&57.7  & 73.4 & 57.2   & 60.4 & 50.5 \\
+$\text{\selftraining{}}^*$~\cite{self-training2005} &  74.2 &	57.4	& 73.4	&57.1 &	60.1& 50.2 \\
+$\text{GradLRE}^*$~\cite{rl_rel} & 74.4& 57.8& 73.3&57.4&60.3&50.5  \\
+${\textsc{\stgg{}}_\textsc{w/o\_AMR}}$ & 74.3 & 58.0 &	73.5 & 57.6 & \textbf{60.5} & 51.1 \\
+${\textsc{\stgg{}}_{\textsc{AMR}}}$ (ours) & \textbf{74.5} 	& \textbf{58.5}  &  \textbf{73.6}  &  \textbf{58.2}   & 60.4 & \textbf{51.7}  \\
\bottomrule
\end{tabular}
}
\end{center}
\vspace{-0.4cm}
\caption{Test F1 scores of event trigger classification (Tri-C), and argument role classification (Arg-C) on three benchmark datasets. 
* denotes methods we re-implement to fit them into the event extraction task. Bold denotes the best performance in each local section and underline denotes the best global performance.
}
\label{tab:main_results}
\vspace{-3mm}
\end{table*}

\subsection{Evaluation of \stgg{} on Event Extraction}

Table~\ref{tab:main_results} shows the event extraction results of both our approach and strong baselines\footnote{We show the variance of Base OneIE and +\stggamr{} on three datasets in Appendix~\ref{app:results}.}. For clarity, in the rest of the section, we refer to our proposed framework as \stggamr{} and our proposed framework with the baseline scoring model as \stggtext{}. We can see that, both \stggamr{} and \stggtext{} improve the performance of the event extraction models on argument role classification while the vanilla \selftraining{} and GradLRE barely work, demonstrating the effectiveness of leveraging the feedback to the pseudo labels during self-training.

We further analyze the reasons in terms of why the vanilla \selftraining{} and GradLRE do not work and notice that: due to the data scarcity, the base event extraction model (i.e., OneIE) performs poorly on many argument roles (lower than 40\% F-score). Thus, the event predictions on unlabeled corpora can be very noisy and inaccurate. The model suffers from confirmation bias~\citep{mean_teacher,confirmation_bias,meta_pseudo-labels}: it accumulates errors and diverges when it's iteratively trained on such noisy pseudo labeled examples during \selftraining{}. 
In addition, we also notice that with \selftraining{}, the event extraction model becomes overconfident about its predictions. We check the averaged probability of all the argument role predictions on the unlabeled dataset which is 0.93. In such case, it is clear that the predicted probability can not faithfully reflect the correctness of the predictions, which is referred as the calibration error~\cite{calibration_error, calibration_error_binary}. 
Thus, the \selftraining{} process which relies on overconfident prediction can become highly biased and diverge from the initial baseline model. 
In GradLRE, the quality of the reward is highly depend on the averaged gradient direction computed during the supervised training process. However, due to the scarcity of the training data, the stored gradient direction can be unreliable. 
In addition, the gradient computed on the pseudo-labeled dataset with high reward is used to update the average gradient direction, which can introduce noises into the reward function. As seen in Table~\ref{tab:main_results}, the best models of \selftraining{} and GradLRE are on par or worse than the baseline approach, and these approaches show the detrimental effects as they show a continuous decline of the performance as training proceeds.

By considering AMR structure, \stggamr{} encourages the event extraction models to predict event structures that are more compatible with AMR graphs. 
This claim is supported by Table~\ref{tab:compatibility_scores}, 
which compares the compatibility scores between the model without \stgg{} (OneIE baseline) and one with \stgg{} (OneIE +\stgg{}) framework on the three benchmark datasets.
The compatibility scores are measured by the AMR based scoring models. We can clearly see that the compatibility scores measured on OneIE+\stggamr{} are much higher than the scores measured on base OneIE. 

\begin{figure*}[ht!]
    \centering
    \includegraphics[width=0.95\linewidth]{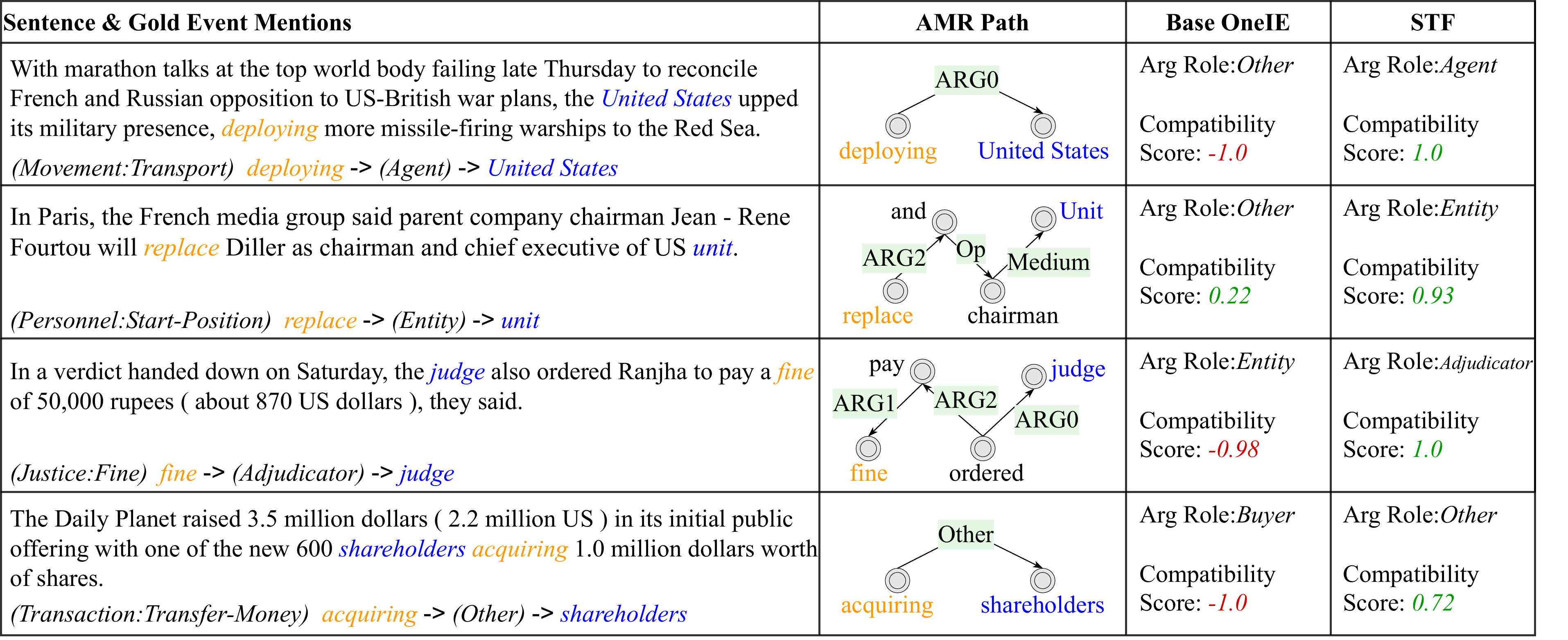}
    \vspace{-2mm}
    \caption{Qualitative results of \stgg{}. Examples are taken from the development and test splits of ACE05-E. The orange tokens denote event triggers and blue tokens denote arguments. The AMR paths are between event triggers and arguments. The Base OneIE and \stgg{} fields show the predicted argument roles from two methods respectively. All the predictions from \stgg{} are correct. The compatibility scores are computed by the same scoring model. Note that OneIE and STF do not use AMR graph at inference time and  AMR graph is shown just to provide intuitions.
    }    
    \label{fig:stf_qualitative_results}
    \vspace{-4mm}
\end{figure*}

Lastly, we observe that OneIE+\stggamr{} outperforms AMR-IE+\stggamr{}, even when AMR-IE performs better than OneIE baseline without \stgg{}.
We argue the reason is that even though \stggamr{} does not need AMR parsing at inference time, AMR-IE does require AMR graphs at inference time which causes it to suffer from potential errors in the AMR parsing. On the other hand, OneIE trained by \stggamr{} does not require AMR graphs at inference time, making it free from potential error propagation. Figure~\ref{fig:stf_qualitative_results} shows more examples to illustrate how the feedback from AMR structures in \stgg{} helps to improve event predictions.



\begin{table}[h]
\begin{center}
\resizebox{\columnwidth}{!}{%
\begin{tabular}{ l | r | r | r | r | r | r }
\toprule
& \multicolumn{2}{c|}{ACE05-E} &
\multicolumn{2}{c|}{ACE05-E+} &
\multicolumn{2}{c}{ERE-EN} \\
\midrule
& Dev & Test & Dev & Test & Dev & Test  \\
\midrule
Base OneIE & 70.1 & 68.4 & 76.9 & 61.9 & 76.4 & 69.2\\
+ ${\textsc{\stgg{}}_{\textsc{AMR}}}$ & \textbf{72.2} & \textbf{70.8} & \textbf{80.2} & \textbf{64.0}& \textbf{78.0} & \textbf{75.1}\\
\bottomrule
\end{tabular}
}
\end{center}
\vspace{-0.4cm}
\caption{The compatibility scores computed by scoring models on the development and test sets of the three benchmark datasets.}
\label{tab:compatibility_scores}
\vspace{-6mm}
\end{table}

\subsection{Effect of Confidence Threshold} \label{data_size}
\begin{figure}[h]
\vspace{-4mm}
    \centering
    \includegraphics[width=0.8\columnwidth]{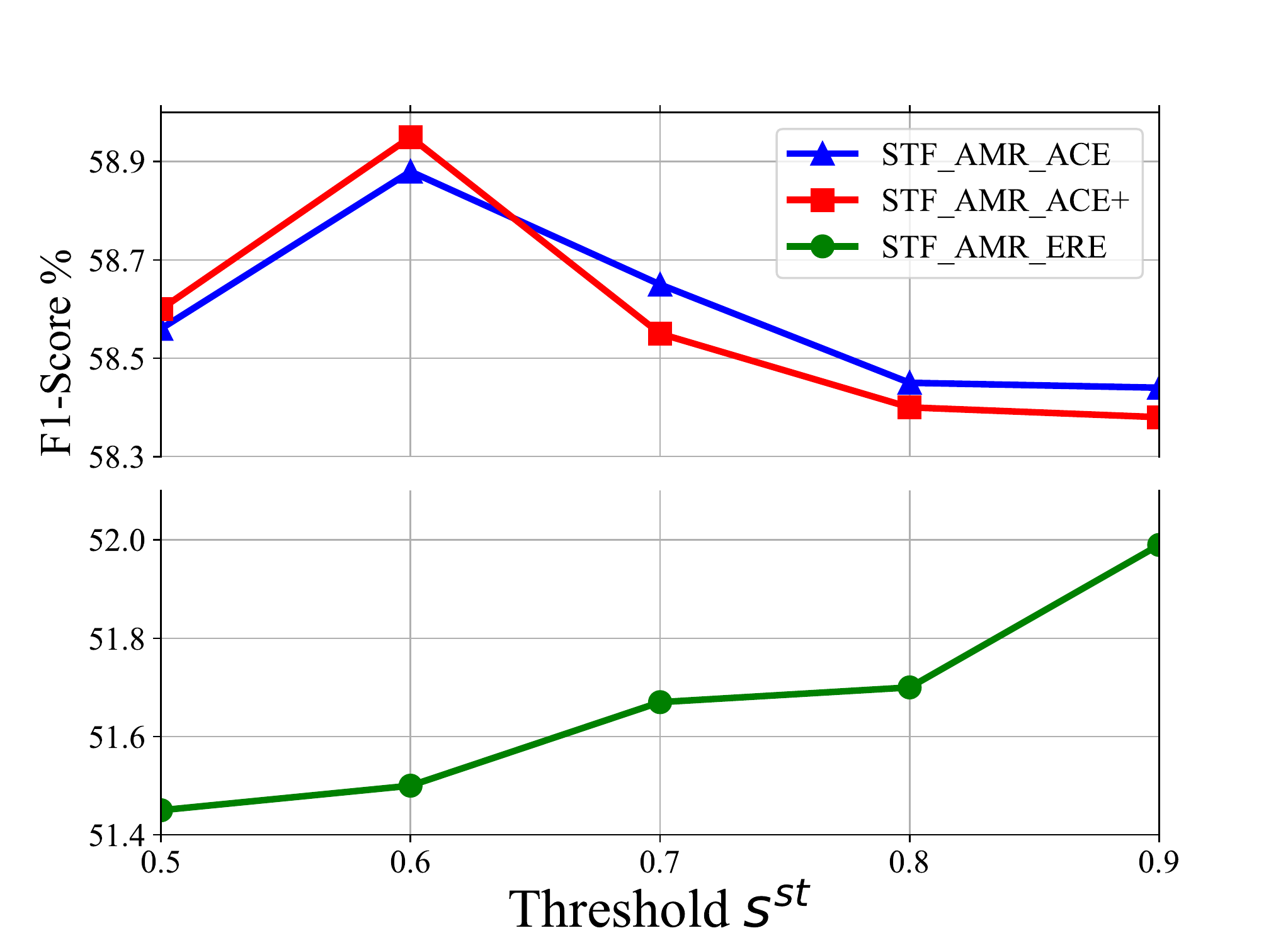}
    \vspace{-2mm}
    \caption{Performance change with different thresholds to select certain pseudo labeled examples for self-training. 
    }
    \label{fig:effect_of_confident_level}
    \vspace{-5mm}
\end{figure}

Intuitively, \stgg{} can leverage both certain (including compatible and incompatible) and uncertain pseudo labeled examples, as when the example is uncertain, the probability $c$ predicted by the scoring model is close to 0.5 and thus $f^c(c)$ is close to $0$, making the gradients computed on this pseudo-labeled example close to $0$. To verify this claim, we conduct experiments with ${\textsc{\stgg{}}_\textsc{AMR}}$ by using the probability $c$ predicted by the scoring model to determine certain and 
uncertain pseudo labels and analyzing their effect to ${\textsc{\stgg{}}_\textsc{AMR}}$. Note that we don't use the probability from the base event extraction model due to its calibration error~\cite{calibration_error}~\footnote{See detailed explanations in Appendix~\ref{sec:self-training_not_work}}. Specifically, we first select a threshold $s^{st}\in\{0.5, 0.6, 0.7, 0.8, 0.9\}$. For each pseudo example, if the probability $c$ predicted by the scoring model is higher than $s^{st}$ (indicating a confident positive prediction) or lower than $1-s^{st}$ (indicating a confident negative prediction), we will add it for ${\textsc{\stgg{}}_\textsc{AMR}}$. The higher the threshold $s^{st}$, the most certain pseudo labels we can select for ${\textsc{\stgg{}}_\textsc{AMR}}$. As Figure~\ref{fig:effect_of_confident_level} shows, ${\textsc{\stgg{}}_\textsc{AMR}}$ can even benefit from the less-confident pseudo labeled examples with threshold $s^{st}$ around 0.6, demonstrating that it can make better use of most of the predicted events from the unlabeled corpus for \selftraining{}.

\subsection{Impact of AMR Parsing}
AMR annotations are very expensive and hard to obtain. To show the potential of \stggamr{} in the scenarios where gold AMR parsing is not available, we conduct experiments by leveraging the NYT 2004 corpus as the external unlabeled corpus with system generated AMR parsing for self-training. 
As shown in Table~\ref{tab:amr_parsing}, with system-based AMR, \stgg{} can also improve the performance of base event extraction models on all three benchmark datasets, and improve over the baseline scoring model without using AMR. The gap between \stgg{} with gold AMR and \stgg{} with system AMR is small, demonstrating that \stgg{} is more robust to the potential errors from AMR parsing.  


\begin{table}[ht!]
\begin{center}
\resizebox{0.9\columnwidth}{!}{%
\begin{tabular}{ l | c | c | c}
\toprule
  &  ACE-E & ACE-E+ & ERE-EN \\
\midrule
Base OneIE & 57.4 & 57.2 & 49.8 \\
+ \stgg{} w/o AMR & 57.9 & 57.6 & 51.0  \\
+ \stgg{} w/ sys\_AMR & 58.2 & 58.1 & 51.4  \\
+ \stgg{} w/ gold\_AMR &58.9  & 59.0 &52.0\\
\bottomrule
\end{tabular}
}
\end{center}
\vspace{-0.4cm}
\caption{Performance comparison between using gold AMR, system-labeled AMR, and not using AMR. 
}
\label{tab:amr_parsing}
\vspace{-4mm}
\end{table}

%% file: 6relatedwork.tex
\section{Related Work}



Most prior studies have been focusing on learning supervised models~\cite{ji2008refining,mcclosky2011event,Li_2013,Chen2015,feng2016language,nguyen2016joint,DyGIE++,Du2020_bert_qa_arg,OneIE,AMR_IE,Sijia2022_Query_Extract,wang2021cleve,FourIE} based on manually annotated event mentions. However, the performance of event extraction has been barely improved in recent years, and one of the main reasons lies in the data scarcity and imbalance of the existing event annotations. Several self-training and semi-supervised studies have been proposed to automatically enrich the event annotations. \newcite{Huang_2012_bootstrap} uses extraction patterns based on nouns that, by definition, play a specific role in an event, to automatically label more data. \newcite{Li_2014_infer} proposes various event inference mechanisms to reveal additional missing event mentions. \cite{huang2020cold,huang2020semi} propose semi-supervised learning to automatically induce new event types and their corresponding event mentions while the performance of old types is also improved. \cite{Liao_2010_bootstrap, Liao_2011_bootstrap, semi_EE} propose techniques to select a more relevant and informative corpus for self-training. All these studies cannot handle the noise introduced by the automatically labeled data properly. Compared with them, our \stgg{} framework leverages a scoring model to estimate the correctness of each pseudo-labeled example, which further guides the gradient learning of the event extraction model, thus it can efficiently mitigate the impact of the noisy pseudo-labeled examples.

Self-training has been studied for many years~\cite{st_word_sense_disambiguation,st_extraction_pattern, self-training2005} and widely adopted in many tasks including speech recognition~\cite{st_speech, noise_st_speech}, biomedical imaging~\cite{you2022mineYourOwnAnatomy, you2022bootstrapping}, parsing~\cite{st_parsing,st_bio_parsing}, and pre-training~\cite{st_pretrain}. Self-Training suffers from inaccurate pseudo labels~\cite{confirmation_bias, confirmation_error, meta_self_rel} especially when the teacher model is trained on insufficient and unbalanced datasets. To address this problem, \cite{meta_pseudo-labels, meta_few-shot_self-training, meta_self_rel} propose to utilize the performance of the student model on the held-out labeled data as a Meta-Learning objective to update the teacher model or improve the pseudo-label generation process.  \newcite{rl_rel} leverage the cosine distance between gradients computed on labeled data and pseudo-labeled data as feedback to guide the self-training process. \cite{ConstrainedSRL, xu2021diora} leverage the span of named entities as constraints to improve semi-supervised semantic role labeling and syntactic parsing, respectively.

%% file: 7conclusion.tex
\section{Conclusion}
\vspace{-5pt}
We propose a self-training with feedback (\stgg{}) framework to overcome the data scarcity issue of the event extract task. The \stgg{} framework estimates the correctness of each pseudo event prediction based on its compatibility with the corresponding AMR structure, and takes the compatibility score as feedback to guide the learning of the event extraction model on each pseudo label during self-training. We conduct experiments on three public benchmark datasets, including \ace{}, \aceplus{}, and ERE, and prove that \stgg{} is effective and general as it can improve any base event extraction models with significant gains. We further demonstrate that \stgg{} can improve event extraction models on large-scale 
 unlabeled corpora even without high-quality AMR annotations. 


%% file: 9limitation.tex
\section*{Limitations}
Our method utilizes the AMR annotations as additional training signals to alleviate the data scarcity problem in the event extraction task. 
In this problem setup, generally speaking, AMR annotations are more expensive than event extraction annotations. 
Nonetheless, in reality, the AMR dataset is much bigger than any existing event extraction dataset, and AMR parsers usually have higher performance than event extraction models. Leveraging existing resources to improve event extraction without requiring additional cost is a feasible and practical direction. Our work has demonstrated the effectiveness of leveraging the feedback from AMR to improve event argument extraction. However, it's still under-explored what additional information and tasks can be leveraged as feedback to improve trigger detection.  

We did not have quantitative results for the alignment between AMR and event graphs. The authors randomly sampled 50 event graphs from ACE05-E and found 41 are aligned with their AMR graphs based on human judgment. In future work, more systematic studies should be conducted to evaluate the alignment.

There is a large gap between the validation and testing datasets in terms of label distribution on \ace{} and \ace+{}. We observe that performance improvement on the validation set sometimes leads to performance decreasing on the test set. Both the validation and test dataset miss certain labels for event trigger types and argument role types. The annotations in the training set, validation set, and test set are scarce and highly unbalanced, which causes the low performance on trained models. We argue that a large-scale more balanced benchmark dataset in the event extraction domain can lead to more solid conclusions and facilitate research. 


%% file: Acknowledgments.tex
\section*{Acknowledgments}
We thank the anonymous reviewers and area chair for their valuable time and constructive comments.
This research is based upon work supported by the Amazon Research Award. Jay-Yoon Lee was supported in part by the New Faculty Startup Fund from Seoul National University.

%% file: 8appendix.tex

\section{Groups of AMR Relations} \label{sec:AMR_rel_groups}

Table~\ref{tab:three_dataset_stat} shows the new categories and labels of AMR relations.

\begin{table}[h]
\begin{center}
\resizebox{\columnwidth}{!}{%
\begin{tabular}{ l | r }
\toprule
 Group Label& AMR Relations \\
\midrule
ARG0 & ARG0 \\
ARG1 & ARG1 \\
ARG2 & ARG2 \\
ARG3 & ARG3 \\
ARG4 & ARG4 \\
Destination & destination \\
Source & source \\
Instrument & instrument \\
Beneficiary & beneficiary \\
Prep roles & role starts with prep \\
Op roles & role start with op \\
Entity role & wiki, name \\
Arg-X role & ARG5, ARG6, ARG7 ARG8, ARG9 \\
Place role & location, path, direction \\
Medium role & manner, poss, medium, topic \\
Modifier role & domain, mod, example \\
Part-whole role & part, consist, subevent, subset \\
\midrule
Time role & \makecell{calendar, century, day, dayperiod, decade,\\ era, month, quarter, season, timezone, \\ weekday, year, year2, time} \\
\midrule
Others & \makecell{purpose, li, quant, polarity, \\condition, extent, degree, snt1, \\snt2, ARG5, snt3, concession, \\ord, unit, mode, value,\\ frequency, polite, age, accompanier, \\snt4, snt10, snt5, snt6, \\snt7, snt8, snt9, snt11,\\ scale, conj-as-if, rel} \\

\bottomrule
\end{tabular}
}
\end{center}
\vspace{-0.4cm}
\caption{The 19 groups of the AMR relations used in our paper.}
\label{tab:three_dataset_stat}
\end{table}

\section{The Statistics of Datasets} \label{sec:dataset_stat}

Table~\ref{tab:three_dataset_stat} shows the statistics of the three public benchmark datasets, including ACE05-E, ACE05-E+ and ERE-EN.

\begin{table*}[h]
\begin{center}
\resizebox{1.5\columnwidth}{!}{%
\begin{tabular}{ l | r | r | r | r | r | r | r | r | r }
\toprule
& \multicolumn{3}{c|}{ACE05-E} &
\multicolumn{3}{c|}{ACE05-E+} &
\multicolumn{3}{c}{ERE-EN} \\
\midrule
& Train & Dev & Test & Train & Dev & Test & Train & Dev & Test  \\
\midrule
\# Sent & 17,172 & 923 & 832 & 19,240 & 902 & 676 & 14,736 & 1,209 & 1,163 \\
\# Entities& 29,006 & 2,451 & 3,017 & 47,525 & 3,422 & 3,673 & 38,864 & 3,320 & 3,291 \\
\# Events& 4,202 & 450 & 403 & 4,419 & 468 & 424 & 6,208 & 525 & 551 \\
\bottomrule
\end{tabular}
}
\end{center}
\caption{The statistics of the three benchmarks used in our paper.}
\label{tab:three_dataset_stat}
\end{table*}

\section{Training Details}
For all experiments, we use Roberta-large as the language model which has 355M parameters. We train all of our models on a single A100 GPU.
\label{appendix_training}
\paragraph{Base OneIE} We follow the same training process as~\cite{OneIE} to train the OneIE model. We use BertAdam as the optimizer and train the model for 80 epochs with 1e-5 as learning rate and weight decay for the language encoder and 1e-3 as learning rate and weight decay for other parameters. The batch size is set to 16. We keep all other hyperparameters the same as \cite{OneIE}. For each dataset we train 3 OneIE models and report the averaged performance. 

\paragraph{Base AMR-IE}
We follow the same training process as~\cite{AMR_IE} to train the AMR-IE model. We use BertAdam as the optimizer and train the model for 80 epochs with 1e-5 as learning rate and weight decay for the language encoder and 1e-3 as learning rate and weight decay for other parameters. The batch size is set to 16. We keep all other hyperparameters exactly the same as \cite{AMR_IE}. For each dataset we train 3 AMR-IE models and report the averaged performance.

\paragraph{Scoring Model} We use BertAdam as the optimizer and train the score model for 60 epochs with 1e-5 as learning rate and weight decay for the language encoder and 1e-4 as learning rate and weight decay for other parameters. The batch size is set to 10. The scoring model contains two self-attention layers. We train 3 scoring models and reported the averaged performance.

\paragraph{Self-Training} For \selftraining{} we use SGD as optimizer and continue to train the converged base OneIE model for 30 epochs with batch size 12, learning rate 1e-4, weight decay for the language encoder as 1e-5, and learning rate 1e-3 and weight decay 5e-5 for all other parameters except the CRF layers and global features which are frozen. For \selftraining{}, we use 0.9 as the threshold to select the confident predictions as pseudo-labeled instances. For all the experiments, we train 3 models and report the averaged performance.

\paragraph{Gradient Imitation Reinforcement Learning}
For GradLRE, we use the BertAdam as the optimizer with batch size 16, learning rate 1e-5 and weight decay 1e-5 for the language encoder and learning rate 1e-3 and weight decay 1e-3 for other parameters to first train OneIE model for 60 epochs. The standard gradient direction vector is computed by averaging the gradient vector on each optimization step. Then following the same training process in the original paper, we perform 10 more epochs of \textit{Gradient Imitation Reinforcement Learning}, and set the threshold for high reward as 0.5. For all the experiments, we train 3 models and report the averaged performance.

\paragraph{Self-Training with Feedback from Abstract Meaning Representation}
For \stgg{}, we first train the OneIE model on the labeled dataset for 10 epochs and continue to train it on the mixture of unlabeled data and labeled dataset for 70 more epochs with batch size 10, learning rate 1e-4, weight decay for the language encoder as 1e-5, and learning rate 1e-3 and weight decay 5e-5 for all other parameters. We leverage a linear scheduler to compute the value for the loss combining ratio $\beta$. The value of $\beta$ is computed as $\frac{\text{epoch}}{70}$. For all the experiments, we train 3 models and report the averaged performance. For model selection, we propose a new method called \textit{Compatibility-Score Based Model Selection} which is discussed in the following paragraph.

\paragraph{Compatibility-Score Based Model Selection} The data scarcity problem not only appears in the training data of ACE-05, ACE-05+ and ERE-EN but appears in the development set. For example, in ACE-05, the development set only contains only 603 labeled argument roles for 22 argument role classes and 7 argument role classes have lees than 10 instances. To alleviate this problem, we propose to leverage part of the large-scale unlabeled dataset as a held-out development set. At the end of each epoch, instead of evaluating the event extraction model on the development set, we run the event extraction model on the unlabeled held-out development set to make event predictions and run the scoring model on the event predictions to compute compatibility scores. We utilize the averaged compatibility scores computed on all instances in the unlabeled held-out development datasets as the model selection criteria. We argue this is another application of the scoring model since its goal is to evaluate the correctness of event predictions. The size of the unlabeled held-out development set is 2,000.

\section{Results of Base OneIE and +\stggamr{}}\label{app:results}
We show the F1 scores of Base OneIE and +\stggamr{} on three benchmark datasets with variances denoted. As one can see that Base OneIE and +\stggamr{} have similar variances on all three datasets except ACE05-E+. We leave how to reduce the variance of argument role classification to future work.
\begin{table}[ht!]
\scriptsize
\begin{center}
\resizebox{\columnwidth}{!}{%
\begin{tabular}{l |  c |  c |  c }
\toprule
& \multicolumn{1}{c}{ACE05-E} &
\multicolumn{1}{c}{ACE05-E+} &
\multicolumn{1}{c}{ERE-EN} \\

\midrule
  & Arg-C & Arg-C  & Arg-C \\
\midrule
Base OneIE & 57.4 $\pm$ 1.23 & 57.2 $\pm$ 0.32 & 49.8 $\pm$ 0.45 \\
 +${\textsc{\stgg{}}_{\textsc{AMR}}}$ (ours) & 58.9 $\pm$ 1.28 & 59.0 $\pm$ 1.03&  52.0 $\pm$ 0.40 \\
\bottomrule
\end{tabular}
}
\end{center}
\vspace{-0.4cm}
\caption{Test F1 scores of argument role classification (Arg-C) on three benchmark datasets.
}
\label{tab:main_results2}
\vspace{-3mm}
\end{table}